\documentclass[letterpaper, 10 pt, journal, twoside]{IEEEtran}
\IEEEoverridecommandlockouts   

\usepackage{graphicx}
\usepackage{xcolor}
\usepackage{tabularx}
\usepackage{booktabs}
\usepackage{pythonhighlight}
\usepackage{lipsum} 
\usepackage[]{mdframed}
\usepackage{xcolor}
\usepackage{amsmath}
\usepackage{footnote}
\usepackage{bm}
\usepackage[ruled,vlined, linesnumbered]{algorithm2e}
\usepackage{amsmath, amssymb}
\usepackage{amsfonts}
\usepackage{algpseudocode}
\usepackage{booktabs}
\usepackage{multirow}
\usepackage{makecell}
\usepackage{romannum}
\usepackage{threeparttable}
\usepackage{hyperref}
\usepackage[caption=false]{subfig}
\usepackage{bm}
\DeclareFontFamily{OT1}{pzc}{}
\DeclareFontShape{OT1}{pzc}{m}{it}{<-> s * [1.10] pzcmi7t}{}
\DeclareMathAlphabet{\mathpzc}{OT1}{pzc}{m}{it}

\graphicspath{{./figs/}}

\bstctlcite{IEEEexample:BSTcontrol}

\title{Towards Deploying VLA without Fine-Tuning: Plug-and-Play Inference-Time VLA Policy Steering via Embodied Evolutionary Diffusion}

\author{Zhuo Li$^{1, 2}$, Junjia Liu$^{1}$, Zhipeng Dong$^{1, 2}$, Tao Teng$^{1}$, Quentin Rouxel$^{1}$,  Darwin Caldwell$^{4}$,~\IEEEmembership{Fellow,~IEEE} \\ and Fei Chen$^{*1,2,3}$,~\IEEEmembership{Senior Member,~IEEE}% <-this % stops a space
\thanks{Manuscript received: November, 17, 2025; Accepted March, 13, 2026. This paper was recommended for publication by Editor Jens Kober upon evaluation of the Associate Editor and Reviewers' comments. This work was supported in part by the Research Grants Council of the
Government of the Hong Kong SAR via the Grant 24209021, 14222722, 14213324, 14211723, C7100-22GF. \textit{(*Corresponding author: Fei Chen.)}}
\thanks{$^{1}$Zhuo Li, Junjia Liu, Zhipeng Dong, Tao Teng, Quentin Rouxel and Fei Chen are with the Collaborative and Versatile Robots (CLOVER) Laboratory, T-Stone Robotics Institute, The Chinese University of Hong Kong, Hong Kong (e-mail: zli@mae.cuhk.edu.hk; jjliu@mae.cuhk.edu.hk; zhipengdong@cuhk.edu.hk; tao.teng@cuhk.edu.hk, quentinrouxel@cuhk.edu.hk, f.chen@ieee.org).}
\thanks{$^{2}$Zhuo Li, Zhipeng Dong and Fei Chen are also with the $\Phi$-Institute for Physical Human Intelligence.}
\thanks{$^{3}$ Fei Chen is also with the Center for Embodied Artificial Intelligence and Computer Vision, Shenzhen Loop Area Institute, Shenzhen, China.}
\thanks{$^{4}$Darwin Caldwell is with the Department of Advanced Robotics, Istituto Italiano di Tecnologia, Genoa, Italy (e-mail: darwin.caldwell@iit.it).}
\thanks{Digital Object Identifier (DOI): see top of this page.}
}

\begin{document}
\maketitle
\pagenumbering{arabic}

\begin{abstract}
Vision-Language-Action (VLA) models have demonstrated significant potential in real-world robotic manipulation. However, pre-trained VLA policies still suffer from substantial performance degradation during downstream deployment. Although fine-tuning can mitigate this issue, its reliance on costly demonstration collection and intensive computation makes it impractical in real-world settings. In this work, we introduce VLA-Pilot, a plug-and-play inference-time policy steering method for zero-shot deployment of pre-trained VLA without any additional fine-tuning or data collection. We evaluate VLA-Pilot on both simulation and real-world experiments across distinct robotic embodiments. Experimental results demonstrate that VLA-Pilot substantially boosts the success rates of off-the-shelf pre-trained VLA policies, enabling robust zero-shot generalization to diverse downstream tasks and embodiments. Experimental videos and code are available at: https://rip4kobe.github.io/vla-pilot/.
\end{abstract}


\section{Introduction}
\IEEEPARstart{R}{ecent} advances in VLA models have substantially improved the generalization capabilities of robotic manipulation \cite{black2024pi_0,liu2025rdtb,wen2025diffusionvla,li2025language,li2025manidp,liu2025human}. By learning from large-scale demonstrations \cite{vuong2023open}, these generalist policies enable robots to acquire a wide repertoire of skills. At inference time, they can perform diverse and contextually appropriate tasks by stochastically sampling actions from the learned skill distribution. Despite these advances, pre-trained VLA policies often suffer performance degradation when deployed on downstream tasks \cite{firoozi2023foundation}. A common approach to mitigate such deployment failures is fine-tuning with task-specific data \cite{kim2025fine}. While effective, this strategy is impractical in real-world applications due to the high cost of data collection and computational resources, as well as the risk of compromising the generalist capabilities of the pre-trained policies. In fact, such deployment failures do not necessarily indicate that the pre-trained VLA policy is incapable of generating the correct behavior. The desired behavior mode may already exist within the policy's generative distribution, but due to suboptimal mode selection at runtime, it fails to be executed reliably \cite{nakamoto2024steering}.

\begin{figure}[t] %H为当前位置，!htb为忽略美学标准，htbp为浮动图形
\centering %图片居中width=0.8\textwidth
\includegraphics[width=0.45\textwidth]{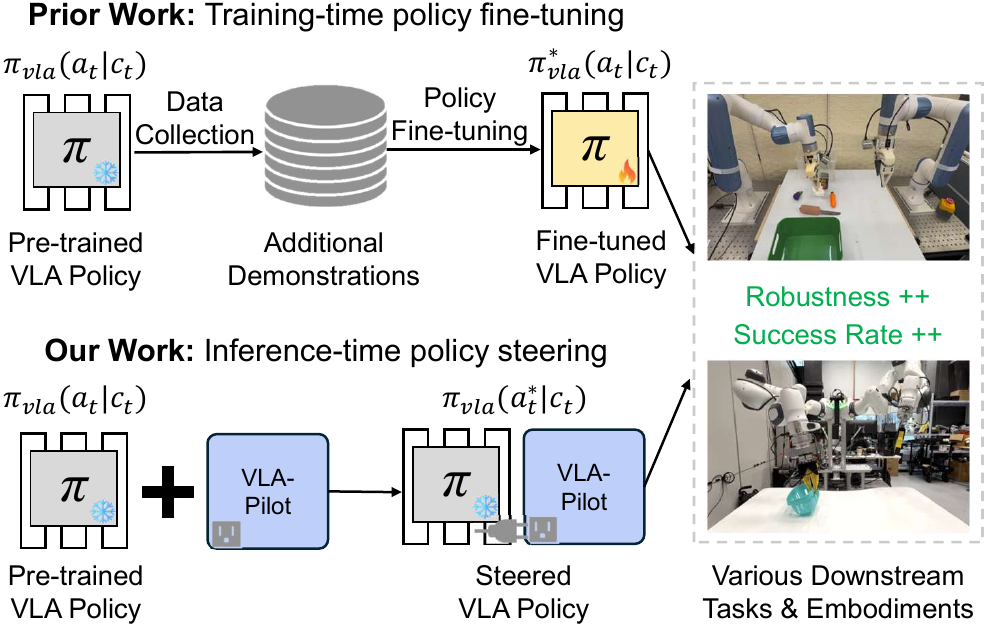} %插入图片，[]中设置图片大小，{}中是图片文件名
\caption{\textbf{Illustration of VLA policy steering.} Prior methods enhance pre-trained VLA policies for downstream tasks through \textit{training-time policy fine-tuning}. In contrast, we propose VLA-Pilot, an \textit{inference-time policy steering} method that enables zero-shot deployment of pre-trained VLA policies without any additional fine-tuning or data collection.} %最终文档中希望显示的图片标题
\label{Fig1} %用于文内引用的标签
\end{figure}

\begin{figure*}[t] %H为当前位置，!htb为忽略美学标准，htbp为浮动图形
\centering %图片居中
\includegraphics[width=0.95\textwidth]{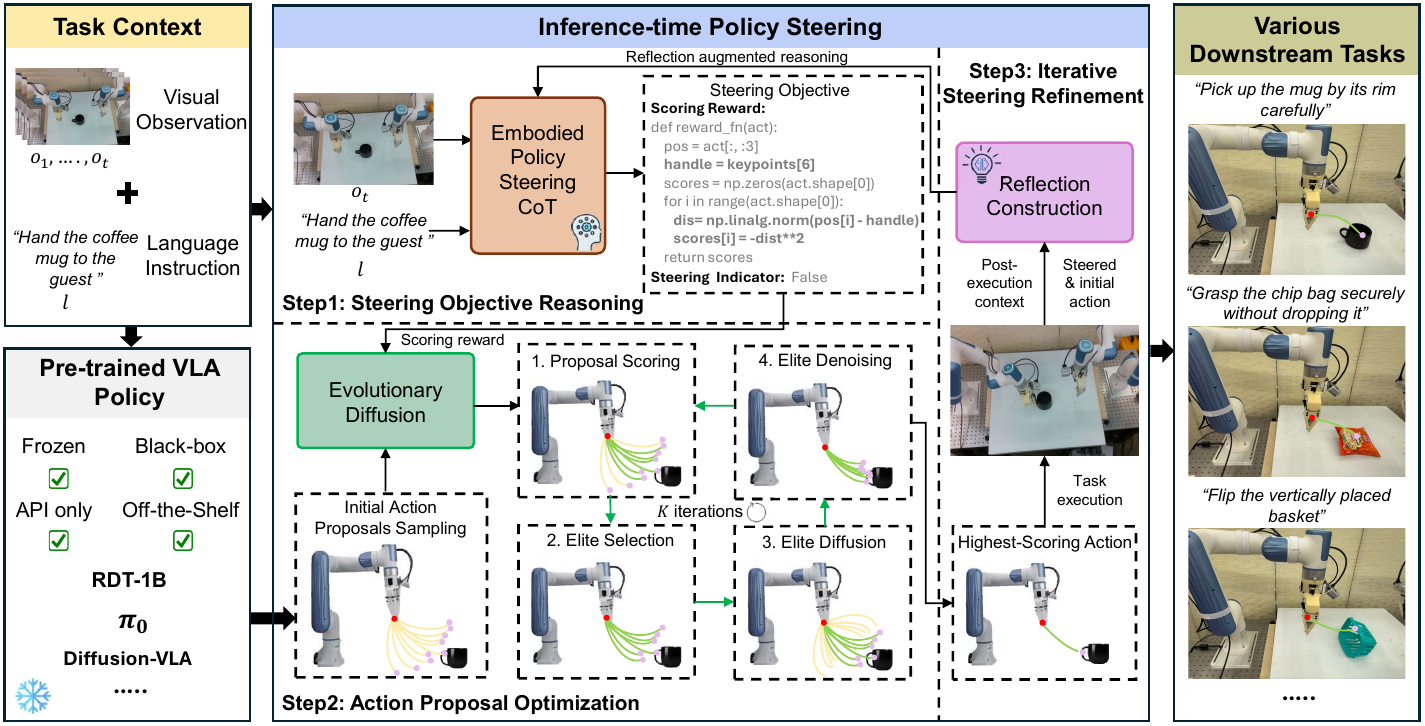}%插入图片，[]中设置图片大小，{}中是图片文件名
\caption{\textbf{Overview of VLA-Pilot.} 
Given a task context, VLA-Pilot steers a pre-trained VLA policy at inference-time via three key steps: 1) \textit{Steering Objective Reasoning} employs EPS-CoT module to reason a task-aligned steering objective reward from the given task context; 2) \textit{Action Proposal Optimization} leverages Evolutionary Diffusion to score and optimize action proposals from the pre-trained VLA based on the reasoned objective reward, and executes the highest-scoring proposal; 3) \textit{Iterative Steering Refinement} integrates post-execution reflection into the EPS-CoT reasoning loop, enabling closed-loop refinement for improved steering accuracy and robustness.}
%最终文档中希望显示的图片标题
\label{Fig2} %用于文内引用的标签
\end{figure*}

Inference-time policy steering \cite{nakamoto2024steering, wagenmaker2025steering, wang2025inference, wu2502foresight, rouxel2025extremumflowmatching} offers an elegant solution to the mode selection problem in pre-trained generative robot policies. By leveraging an external verifier to evaluate and select task-aligned candidate actions proposed by the pre-trained policy, robot behavior can be effectively \textit{steered} at runtime without the need for policy fine-tuning. However, existing methods face two key limitations. First, the verifiers used in these approaches typically require additional training and often exhibit limited generalization due to the narrow distribution of their training data \cite{nakamoto2024steering}. Second, these methods rely solely on selecting actions from a fixed set of proposals \cite{wu2502foresight, nakamoto2024steering}. However, in complex downstream tasks, the pre-trained VLA policy may fail to generate any candidate action that aligns with the task context. In such cases, the verifier cannot recover a successful behavior through selection alone, leading to steering failures during deployment.

To address these limitations, we propose VLA-Pilot, a training-free inference-time policy steering method that improves both the generalization and task alignment of pre-trained VLA policies for downstream deployment (as shown in Figure \ref{Fig1}). Our core idea is to leverage Multimodal Large Language Model (MLLMs) as open-world verifiers to enhance generalization, and to employ an Evolutionary Diffusion process as an action optimizer to improve task alignment. Specifically, given a downstream task context, VLA-Pilot first employs an Embodied Policy Steering Chain-of-Thought (EPS-CoT) module to infer steering objectives reward using the open-world reasoning capabilities of MLLMs. This removes the need for training task-specific verifiers and significantly improves generalization to out-of-distribution tasks. Next, VLA-Pilot introduces a novel Evolutionary Diffusion algorithm to optimize action proposals sampled from the pre-trained VLA policies. Unlike previous selection-based steering method, Evolutionary Diffusion not only selects, but also evolves action candidates toward a task-aligned distribution, enabling effective policy steering even when initial proposals are suboptimal or infeasible. Finally, VLA-Pilot incorporates an iterative steering refinement mechanism to perform closed-loop correction, enhancing steering accuracy and robustness. An overview of VLA-Pilot is shown in Figure \ref{Fig2}. In summary, we explore a promising paradigm that focuses on maximizing the utility of existing VLA models during inference, rather than pursing increasingly larger datasets and architectures. We demonstrate that the pre-trained VLA models already encapsulate sufficient latent knowledge to solve new tasks, and that such knowledge can be effectively extracted and aligned with task objectives through the proposed steering mechanism. Our contributions are threefold:

\begin{itemize}
    \item {VLA-Pilot, a plug-and-play inference-time policy steering method that enables zero-shot generalization of frozen VLA policies across diverse downstream tasks and embodiments， without requiring any additional policy fine-tuning or data collection.}
    \item {An Embodied Reasoning Guided Evolutionary Diffusion strategy that jointly infers generalized steering objectives and optimizes action proposals for enhanced task alignment.}
    \item {We conduct extensive experiments across six real-world manipulation tasks and two distinct robotic embodiments. Results show that VLA-Pilot improves the average success rate of two pre-trained VLA policies by 31$\%$, significantly outperforming all baseline methods.}
\end{itemize}

\section{Related Work}
\subsection{Inference-Time Policy Steering}
Inference-time policy steering has emerged as a promising approach to enhance generalist VLA policies without fine-tuning \cite{nakamoto2024steering, wagenmaker2025steering, wang2025inference,wu2502foresight, dai2025rover, song2025hume, kwok2025robomonkey}. It assumes the pre-trained policy can generate viable actions but struggles with selection. To address this, external verifiers such as human feedback \cite{wang2025inference}, Q-functions \cite{nakamoto2024steering, song2025hume}, or VLMs \cite{wu2502foresight, kwok2025robomonkey} are used to score and choose among action proposals. Our work builds upon the use of Foundation Models (FMs) as verifiers \cite{wu2502foresight}, while introducing two key innovations. First, instead of explicitly scoring proposals, we leverage FMs reasoning to infer implicit task goals and synthesize a high-level reward objective. Second, we go beyond selecting from fixed proposals by introducing an Evolutionary Diffusion algorithm that refines the proposal distribution toward the inferred objective. This shift from static action selection to dynamic action optimization improves steering robustness and generalization.

\subsection{Foundation Models for Robot Manipulation}
Recent advances in Foundation Models \cite{firoozi2023foundation} have significantly advanced robotic manipulation by providing generalizable semantic priors. Existing studies primarily explore two distinct avenues: high-level task planning \cite{Huang2022InnerME, lin2023text2motion} and low-level skill generation via reward synthesis \cite{HuangWLZF24, Ma2023EurekaHR,zhang2024grape}. While our method aligns more closely with the latter, it introduces a key distinction in how the synthesized reward is utilized. Prior works either employ derived rewards to supervise policy training \cite{Ma2023EurekaHR,zhang2024grape}, or as direct control signals at execution time \cite{HuangWLZF24}. In contrast, we propose a novel inference-time policy steering mechanism, wherein the synthesized reward is used to guide a frozen VLA policy at deployment via online reward-conditioned action optimization. This enables flexible, zero-shot adaptation to novel task variations without requiring policy updates or additional interaction data.

\section{Methods}\label{METHODS}

\subsection{Problem Formulation} \label{section:gd}
Given a downstream task context \( c_t = (o_t, l) \) that contains visual observation \( o_t \) and language instruction \( l \), we study the policy steering problem for a pre-trained diffusion-based VLA policy \( \pi_{\text{vla}}(a_t \mid c_t) \). The objective is to identify an action proposal $a_t^\star$ that best aligns with $c_t$ at runtime:
\begin{equation}
    a_t^\star = \arg\max_{a_t \in A^0} R(a_t; c_t),
    \label{eq1}
\end{equation}
where $A^0=\{a_t^i\sim\pi_{\text{vla}}(a_t \mid c_t)\}_{i=1}^M$ is $M$ i.i.d. action proposals sampled from the pre-trained VLA policy, and \( R(a_t; c_t) \) denotes a steering objective reward that measures how well action \( a_t \) aligns with the task context \( c_t \). Solving the policy steering problem defined in Eq.~\ref{eq1} requires two key capabilities: steering objective reasoning and action proposal optimization. The first involves inferring a task-aligned objective reward \( R(a_t; c_t) \) from the context \( c_t \). VLA-Pilot addresses this via the EPS-CoT module, which leverages MLLMs to reason about open-world steering objectives. The second capability involves searching within the action distribution of the pre-trained VLA policy to identify feasible actions. VLA-Pilot addresses this via Evolutionary Diffusion, which iteratively scores and mutates sampled action candidates based on the steering reward  \( R(a_t; c_t) \), enabling effective adaptation to downstream task requirements.

\subsection{Steering Objective Reasoning}
The first step in policy steering is to infer a task-aligned steering objective reward \( R(a_t; c_t) \) from the given context $c_t$. Our key insight is that the policy steering problem in VLA bears a strong analogy to the prompting problem in LLMs. Therefore, inspired by the effectiveness of Chain-of-Thought (CoT) \cite{wei2022chain}, we propose Embodied Policy Steering Chain-of-Though, a structured reasoning module designed to generate steering objective rewards:
\begin{equation}
   R(a_t; c_t) = \mathcal{F}_{\text{EPS-CoT}}(\Phi_{\text{MLLM}}(c_t)).
\end{equation}

\begin{figure}[!t]
\centering
\includegraphics[width=0.9\linewidth]{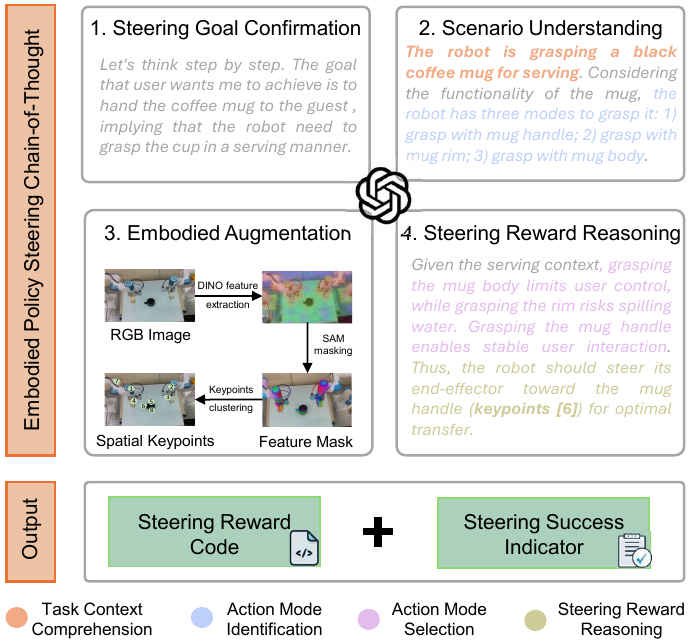}
\caption{\textbf{Embodied Policy Steering Chain-of-Thought.} EPS-CoT guides the steering objective reasoning process through a structured CoT.}
\label{fig3}
\end{figure}

As illustrated in Figure \ref{fig3}, EPS-CoT decomposes the reasoning process into four interleaved stages. It begins with \textit{steering goal confirmation}, where the MLLM is prompted to rephrase and verify the language instruction in order to ensure alignment between the task demand and the steering goal. Following this, the MLLM performs \textit{scenario understanding}, where it interprets the task context and identifies potential action modes based on the visual observation. This step facilitates a high-level understanding of the task scenario, including environmental affordances, spatial relationships, and task-relevant entities. To further ground embodied information in the reasoning process , EPS-CoT incorporates \textit{embodied augmentation} \cite{zawalski2024robotic}, which enhances the reasoning by integrating spatial keypoints of robot end-effector position and object location extracted via vision foundation models, namely DINO \cite{jose2025dinov2} and SAM \cite{ravi2024sam}. Finally, based on scenario understanding and embodied information, EPS-CoT infers the task-aligned steering objective and generates the corresponding scoring reward code. Given the inherent ambiguity and imprecision of natural language instructions, we implement the reward as a non-differentiable black-box scoring function \cite{yang2024diffusion}. This formulation effectively captures the vague yet goal-directed nature of language, while also simplifying the reasoning process required by the MLLM.

\subsection{Action Proposal Optimization}
To improve task alignment of initial action proposals, we introduce an Evolutionary Diffusion algorithm that synergistically leverages the multimodal expressiveness of diffusion process and the black-box optimization of evolutionary search~\cite{zhang2024diffusion}. The proposed algorithm begins by sampling initial action proposals $A^0$ of $M$ actions using the pre-trained VLA policy:
\begin{equation}
   A^0 = \{a_t^i \sim \pi_{\text{vla}}(a_t \mid c_t) \mid i = 1, \ldots, M\}.
\end{equation}

\begin{figure}[!t]
\centering
\includegraphics[width=0.9\linewidth]{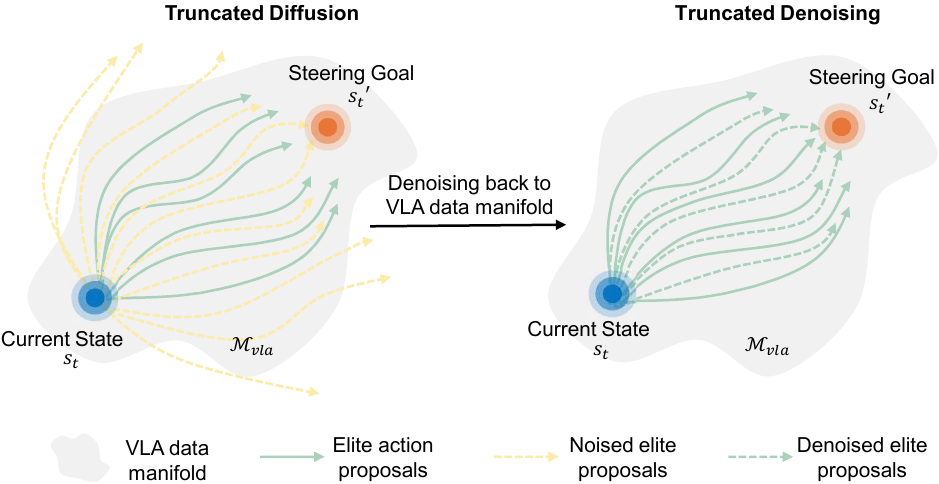}
\caption{\textbf{Truncated Diffusion-Denoising Process.} VLA-Pilot employs a truncated diffusion-denoising mechanism to mutate elite proposals, thereby enhancing action diversity and exploration capabilities to achieve better task alignment.}
\label{fig4}
\end{figure}

We then perform an evolutionary search loop to iteratively evaluate and mutate the initial proposal set using the steering objective reward $R(a_t; c_t)$. Specifically, at each evolutionary iteration $k$, we score the proposal set $\{R(a_t^i; c_t)|a_t^i\in A^{k-1} \}_{i=1}^M$ and select high-scoring elite proposals $E^{k}\subseteq A^{k-1}$:

\begin{equation}
q(a_t) = \frac{\exp\left( \tau R(a_t; c_t) \right)}
             {\sum_{i=1}^{M} \exp\left( \tau R(a_t^i; c_t) \right)}
\end{equation}
\begin{equation}
E^{k} = \left\{ a_t^i \overset{\text{iid}}{\sim} q(a_t) \right\}_{i=1}^{M},
\end{equation}
where $\tau$ is a tunable temperature parameter controlling the sharpness of $q$. To further enhance proposal diversity and explore task-aligned actions, we apply a truncated diffusion-denoising process to mutate elite proposals $E^{k}$ (see Figure \ref{fig4}). Specifically, we run the first $n$ steps of the forward diffusion process to obtain noised elite proposals $\bar{E}^{k}$:
\begin{equation}
\bar{E}^{k} = \left\{ \sqrt{\bar{\alpha}_N} a_t + \sqrt{1 - \bar{\alpha}_N} \, \epsilon \,\middle|\, a_t \in E^{k} \right\},
\end{equation}
where $\epsilon \sim \mathcal{N}(0, 1)$, $\bar{\alpha_N}$ denotes the cumulative product of noise schedule coefficients up to diffusion step $N$. However, directly applying forward diffusion may lead to $\bar{E}^{k}$ drifting out of the original VLA distribution due to the stochastic nature of the noise. Therefore, we subsequently perform the final $n$ steps of the reverse diffusion process, using the noise predictor from the pre-trained VLA policy to denoise $\bar{E}^{k}$ and obtain the refined proposals $A^{k}$ for propagation, ensuring they lie within the original data manifold:
\begin{equation}
A^{k} = \left\{ \bar{a_t} \sim \pi_{\text{vla}}(\bar{a_t} \mid c_t) \,\middle|\, \bar{a_t} \in \bar{E}^{k} \right\}.
\end{equation}

Finally, after the evolutionary search loop is completed, we select the highest-scoring elite action that best aligns with the steering objective for execution.

\subsection{Iterative Steering Refinement}
We introduce an Iterative Steering Refinement mechanism that facilitates closed-loop correction of both the steering objective and the resulting actions. Concretely, we augment the original EPS-CoT reasoning module with a reflection step \cite{Ma2023EurekaHR}, in which the MLLM is prompted using four key components: the robot initial action $a_0$ before the steering process, the selected execution proposal $a_t^\star$, the updated post-execution task context $\bar{c}_t$, and the reasoning history $\mathcal{H}_t$ from the preceding EPS-CoT step. Given this reflection-augmented input, the MLLM serves as a self-critic to refine the steering reward $R(a_t^i; c_t)$ and produce the steering success indicator $s$:
\begin{equation}
    s = \mathcal{F}_{\text{EPS-CoT}}(\Phi_{\text{MLLM}}(a_0, a_t^\star, \bar{c}_t, \mathcal{H}_t)).
\end{equation}

If the MLLM detects inconsistencies in the inferred steering reward, a new reward is regenerated. Similarly, if the executed action is misaligned with the task context (i.e., $s=False$), VLA-Pilot continues the refinement process until a maximum number of retries is reached. This closed-loop refinement ensures improvement in both the accuracy and contextual relevance of the steering process. The complete workflow of our proposed VLA-Pilot is summarized in Algorithm \ref{alg:alg1}.

\begin{algorithm}[ht]
\caption{Inference-Time VLA Policy Steering}
\label{alg:alg1}
\KwIn{Pre-trained VLA $\pi_{\text{vla}}(a_t \mid c_t)$, task context $c_t$, MLLM $\Phi_{\text{MLLM}}$, max retries $N_{\text{max}}$}
\KwOut{Task-aligned action $a_t^\star$}

% $n_{\text{retry}} \gets 0$\;
\tcp{Steering objective reasoning}
$R(a_t; c_t) \gets \mathcal{F}_{\text{EPS-CoT}}(\Phi_{\text{MLLM}}(c_t))$\;
\tcp{Action proposal optimization}
Initial proposal sampling: $A^0 = \{a_t^i \sim \pi_{\text{vla}}(a_t \mid c_t) \mid i = 1, \ldots, M\}$\;

\For{$k = 1$ \KwTo $K$}{
    Proposal scoring: $\{R(a_t^i; c_t)|a_t^i\in A^{k-1} \}_{i=1}^M$\;
    Compute score distribution: $q(a_t) = \frac{\exp\left( \tau R(a_t; c_t) \right)}
             {\sum_{i=1}^{M} \exp\left( \tau R(a_t^i; c_t) \right)}$\;
    Elite selection: $E^{k} = \left\{ a_t^i \overset{\text{iid}}{\sim} q(a_t) \right\}_{i=1}^{M}$\;
    Elite diffusion: $\bar{E}^{k} = \left\{ \sqrt{\bar{\alpha}_N} a_t + \sqrt{1 - \bar{\alpha}_N} \, \epsilon \,\middle|\, a_t \in E^{k} \right\}$\;
    Elite denoising: $A^{k} = \left\{ a_t \sim \pi_{\text{vla}}(a_t \mid c_t) \,\middle|\, a_t\in \bar{E}^{k} \right\}$\;
}
$a_t^\star \gets \arg\max_{a_t \in A^K} R(a_t; c_t)$\;
\tcp{Iterative steering refinement}
$\bar{c}_t, \mathcal{H}_t \gets \Call{Execute}{a_t^\star}$\;
$s \gets \mathcal{F}_{\text{EPS-CoT}}(\Phi_{\text{MLLM}}(a_0, a_t^\star, \bar{c}_t, \mathcal{H}_t))$\;
\lIf{$s = \text{True}$}{\KwRet success}
\Else{
    % $n_{\text{retry}} \gets n_{\text{retry}} + 1$\;
    \lIf{retry count $< N_{\text{max}}$}{restart from step 1}
    \lElse{\KwRet failure}
}
\end{algorithm}

\section{Experiments}
We perform extensive experiments in both simulation and real-world to investigate the following questions: 1) Can VLA-Pilot improve the downstream performance of off-the-shelf, pre-trained VLA policies? 2) How does its performance compare to state-of-the-art policy steering baselines? 3) How does VLA-Pilot compare to direct fine-tuning of the underlying VLA policy? 4) Can VLA-Pilot enable cross-embodiment generalization? 5) What are the individual contributions of its core components? 6) What is the inference-time latency introduced by VLA-Pilot?

\begin{table*}[!t]
\caption{Simulation Results on RoboTwin and ManiSkill3 Benchmarks}
\label{tab:sim_results}
\centering
\begin{threeparttable}
\begin{tabular*}{\textwidth}{@{\extracolsep{\fill}} lccccccc}
\toprule
& \multicolumn{4}{c}{\textbf{ManiSkill}} & \multicolumn{2}{c}{\textbf{RoboTwin}} & \\
\cmidrule(lr){2-5} \cmidrule(lr){6-7}
\textbf{Method} & \textbf{PegInsertion} & \textbf{PickCube} & \textbf{PlugCharger} & \textbf{StackCube} & \textbf{LiftPot} & \textbf{DumpBin} & Avg.MSR($\uparrow$) \\
\midrule
ReKep & 0.20 \scriptsize$\pm$ 0.04 & 0.80 \scriptsize$\pm$ 0.06 & 0.04 \scriptsize$\pm$ 0.00 & 0.80 \scriptsize$\pm$ 0.05 & 0.79 \scriptsize$\pm$ 0.03 & 0.70 \scriptsize$\pm$ 0.04 & 0.56 \scriptsize$\pm$ 0.03 \\
RDT-1B & 0.13 \scriptsize$\pm$ 0.05 & 0.77 \scriptsize$\pm$ 0.07 & 0.01 \scriptsize$\pm$ 0.02 & 0.74 \scriptsize$\pm$ 0.06 & 0.72 \scriptsize$\pm$ 0.04 & 0.64 \scriptsize$\pm$ 0.05 & 0.50 \scriptsize$\pm$ 0.04 \\
RDT-1B+ours & 0.24 \scriptsize$\pm$ 0.05 & 0.92 \scriptsize$\pm$ 0.09 & 0.11 \scriptsize$\pm$ 0.04 & 0.91 \scriptsize$\pm$ 0.06 & 0.92 \scriptsize$\pm$ 0.08 & 0.79 \scriptsize$\pm$ 0.05 & \textbf{0.65 \scriptsize$\pm$ 0.06} \\
w/o EPS-CoT & 0.17 \scriptsize$\pm$ 0.06 & 0.83 \scriptsize$\pm$ 0.08 & 0.05 \scriptsize$\pm$ 0.03 & 0.84 \scriptsize$\pm$ 0.07 & 0.83 \scriptsize$\pm$ 0.05 & 0.71 \scriptsize$\pm$ 0.06 & 0.57 \scriptsize$\pm$ 0.06 \\
w/o Evolutionary Diffusion & 0.14 \scriptsize$\pm$ 0.03 & 0.81 \scriptsize$\pm$ 0.05 & 0.02 \scriptsize$\pm$ 0.01 & 0.80 \scriptsize$\pm$ 0.04 & 0.80 \scriptsize$\pm$ 0.06 & 0.66 \scriptsize$\pm$ 0.07 & 0.54 \scriptsize$\pm$ 0.05 \\
w/o Iterative Refinement & 0.18 \scriptsize$\pm$ 0.04 & 0.90 \scriptsize$\pm$ 0.06 & 0.04 \scriptsize$\pm$ 0.02 & 0.89 \scriptsize$\pm$ 0.05 & 0.87 \scriptsize$\pm$ 0.07 & 0.69 \scriptsize$\pm$ 0.04 & 0.60 \scriptsize$\pm$ 0.05 \\
\bottomrule
\end{tabular*}
\vspace{3pt}
\scriptsize
\noindent \textit{Each method was evaluated in a standardized setting using 250 episodes per task (10 training seeds × 25 episodes per seed). We report the avergae MSR and the standard deviation across different seeds.}
\end{threeparttable}
\end{table*}

\begin{table*}[!t]
\caption{Real-robot experiment results across six downstream tasks}
\label{tab1}
\centering
\begin{threeparttable}
\resizebox{\textwidth}{!}{%
\begin{tabular*}{\textwidth}{@{\extracolsep{\fill}} @{}cccccccccccc@{}}
\toprule
\multirow{3}{*}{\centering \textbf{\makecell{Tasks}}} 
& \makecell{\textbf{DiVLA}} 
& \makecell{\textbf{RDT-1B}} 
& \multicolumn{2}{c}{\textbf{DiVLA+V-GPS}}                     
& \multicolumn{2}{c}{\textbf{DiVLA+FOREWARN}}                              
& \multicolumn{2}{c}{\textbf{DiVLA+Ours}}                         
& \multicolumn{2}{c}{\textbf{RDT-1B+Ours}}                         \\ 
\cmidrule(lr){2-2} \cmidrule(lr){3-3} \cmidrule(lr){4-5} \cmidrule(lr){6-7} \cmidrule(lr){8-9} \cmidrule(lr){10-11}
 & \makecell{Avg.\\MSR($\uparrow$)} & \makecell{Avg.\\MSR($\uparrow$)} & \makecell{Avg.\\MSR($\uparrow$)} & \makecell{Avg.\\SOA($\uparrow$)} & \makecell{Avg.\\MSR($\uparrow$)} & \makecell{Avg.\\SOA($\uparrow$)} & \makecell{Avg.\\MSR($\uparrow$)} & \makecell{Avg.\\SOA($\uparrow$)} & \makecell{Avg.\\MSR($\uparrow$)} & \makecell{Avg.\\SOA($\uparrow$)}
\\
\midrule
Mug Handling            & 0.54 & 0.51 & 0.68 & 0.83 & 0.71 & 0.85 & 0.75 & 0.88 & 0.74 & 0.85 \\
Bag Handling            & 0.51 & 0.51 & 0.66 & 0.75 & 0.69 & 0.77 & 0.72 & 0.83 & 0.73 & 0.78 \\
Basket Flipping           & 0.42 & 0.40 & 0.48 & 0.60 & 0.52 & 0.65 & 0.67 & 0.73 & 0.65 & 0.70 \\
Table Bussing       & 0.17 & 0.16 & 0.21 & 0.38 & 0.30 & 0.42 & 0.63 & 0.72 & 0.63 & 0.68 \\
Bimanual Bussing    & 0.11 & 0.12 & 0.13 & 0.32 & 0.21 & 0.34 & 0.58 & 0.67 & 0.55 & 0.63 \\
Bimanual Zippering  & 0.08 & 0.07 & 0.10 & 0.15 & 0.13 & 0.28 & 0.39 & 0.55 & 0.31 & 0.53 \\
\midrule
\textbf{Overall}    & 0.31 & 0.30 & 0.38 & 0.51 & 0.43 & 0.55 & \textbf{0.62} & \textbf{0.73} & 0.60 & 0.69 \\
\bottomrule
\end{tabular*}%
}
\vspace{3pt}
\scriptsize
% \textit{For each method and task, we perform 20 trials and report the mean performance.}
\end{threeparttable}
\end{table*}

\subsection{Setup}
\textit{1) Implementation details:} 
We instantiate the EPS-CoT using GPT-4o with a temperature of 0.2 and a maximum output length of 1000 tokens. For evolutionary diffusion, we sample an initial population of 32 action proposals and perform 10 steps of evolutionary search to iteratively refine actions, balancing search diversity and computational efficiency during inference.

\begin{figure*}[t] %H为当前位置，!htb为忽略美学标准，htbp为浮动图形
\centering %图片居中
\includegraphics[width=0.9\textwidth]{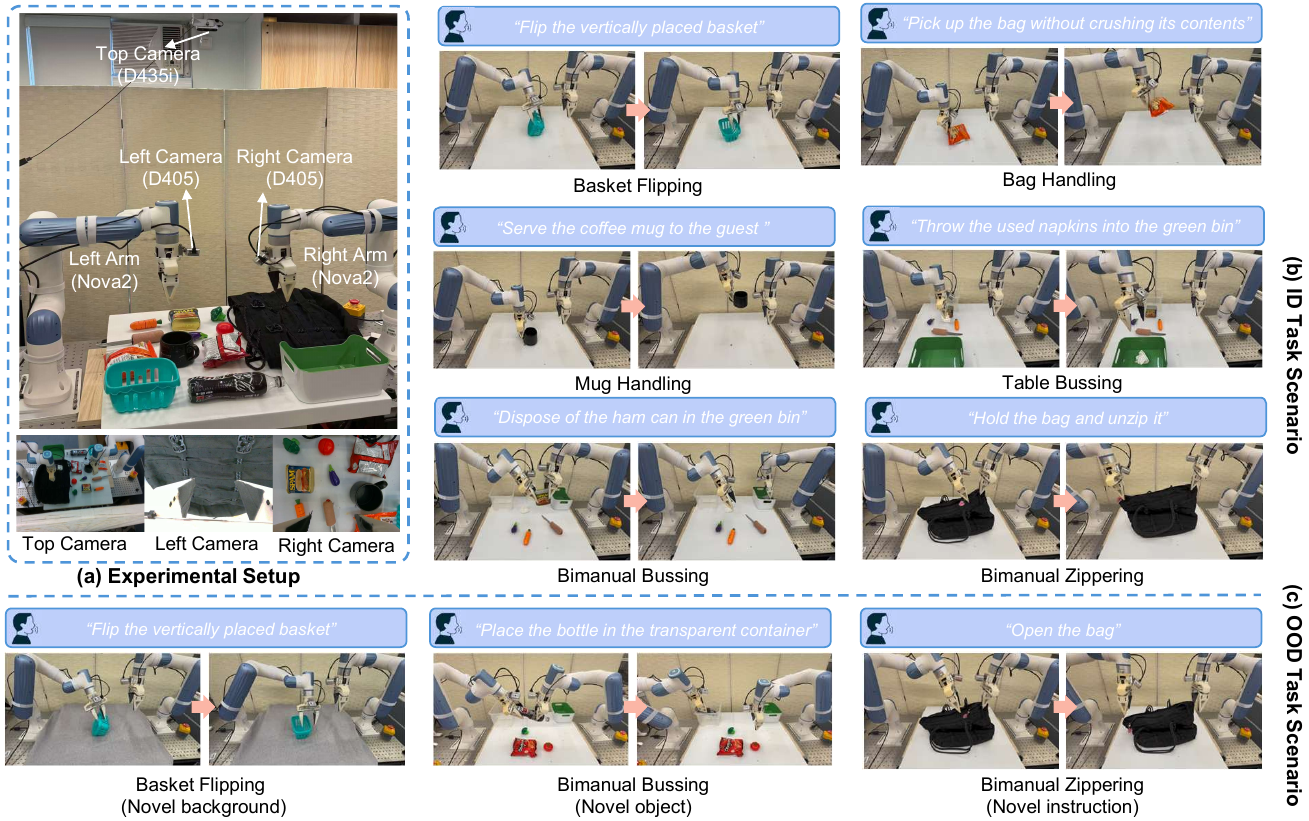}%插入图片，[]中设置图片大小，{}中是图片文件名
\caption{\textbf{Qualitative results of real robot experiments.} 
VLA-Pilot effectively steers off-the-shelf pre-trained VLA policies to complete downstream tasks at inference time, achieving zero-shot deployment across both ID and OOD task scenarios.}
%最终文档中希望显示的图片标题
\label{Fig5} %用于文内引用的标签
\end{figure*}

\textit{2) Task settings:} 
We adopt two widely used simulation benchmarks for evaluation: RoboTwin~\cite{chen2025robotwin} and ManiSkill3~\cite{tao2024maniskill3}. For single-arm settings, we evaluate on four representative tasks from ManiSkill3: PickCube, StackCube, PlugCharger, and PegInsertionSide. For dual-arm settings, we focus on two representative tasks from RoboTwin: LiftPot and DumpBin. Real-world experiments were conducted using the dual-arm DOBOT X-Trainer system, which consists of two 6-DoF Nova2 manipulators, each equipped with a 1-DoF gripper. The setup includes three Intel RealSense cameras for capturing RGB image observations (see Figure~\ref{Fig5}(a)). The evaluation covers six downstream tasks: four relatively simple single-arm manipulations and two more complex dual-arm tasks (Figure~\ref{Fig5}(b)). To assess generalization capabilities, we designed two types of task scenarios: In-Distribution (ID) and Out-of-Distribution (OOD), depending on whether the scenario had been encountered by the external verifier used in baseline steering methods during training. For each scenario, five task-specific language instructions were provided to evaluate the performance of each method.

\textit{4) Evaluation metrics:}
Two quantitative metrics are adopted for the experiments. \textbf{Manipulation Success Rate (MSR)} measures the proportion of trials that successfully complete the downstream task. \textbf{Steering Objective Alignment (SOA)} quantifies the proportion of trials where the post-execution state falls within a predefined threshold of the target keypoint, indicating alignment with the intended steering objective.

\textit{5) Baseline methods:} We evaluate VLA-Pilot against the following baselines: 
\begin{itemize}
    \item \textbf{Diffusion-VLA (DiVLA) \cite{wen2025diffusionvla},} a 2B parameter pre-trained VLA policy that integrates autoregression with a diffusion model;
    \item \textbf{RDT-1B \cite{liu2025rdtb},} a diffusion-based VLA policy for generalized robotic manipulation;
    \item \textbf{V-GPS \cite{nakamoto2024steering},} an inference-time VLA policy steering method that selects optimal action using a trained value function-based verifier; 
    \item \textbf{FOREWARN \cite{wu2502foresight},} a VLM-in-the-loop policy steering method that combines a World Model with a fine-tuned VLM verifier for action ranking. For each task, we collected 100 demonstrations and 200 VLA policy rollouts to train the World Model and fine-tune the VLM verifier.
    \item \textbf{DiVLA-finetune}, a DiVLA policy fine-tuned on 50 task demonstrations;
    \item \textbf{RDT-1B-finetune}, an RDT-1B policy fine-tuned on 50 task demonstrations.
    \item \textbf{ReKep \cite{HuangWLZF24}}, a VLM-guided manipulation baseline that leverages keypoint constraints for trajectory optimization.
\end{itemize}

\subsection{Results}
\textbf{VLA-Pilot Improves Pre-trained VLA Policy.} 
Across all six simulation tasks, VLA-Pilot consistently outperforms its VLA baselines, achieving an average MSR of 0.65, which is 0.15 higher than the base RDT-1B policy (Table~\ref{tab:sim_results}). In real-world experiments (Table~\ref{tab1}), VLA-Pilot improves both DiVLA and RDT-1B across all six downstream tasks, with average MSR gains of +0.31 and +0.30, respectively. Qualitative results (Figure~\ref{Fig5}) further support these findings, demonstrating that the integration of VLA-Pilot facilitates zero-shot deployment of pre-trained VLA policies to downstream tasks. We observe that failure cases of VLA baselines primarily stem not from action infeasibility, but from inconsistent selection of task-relevant actions. For example, in the Mug Handling task, which involves diverse manipulation modes (e.g., grasping, lifting, pouring), DiVLA achieves an MSR of 0.54, indicating that valid actions can be generated. However, due to its inability to consistently select actions aligned with the task semantics, its final MSR is significantly lower than the version augmented with VLA-Pilot, which achieves a much higher MSR of 0.75.

 \begin{figure}[!t]
\centering
\includegraphics[width=0.9\linewidth]{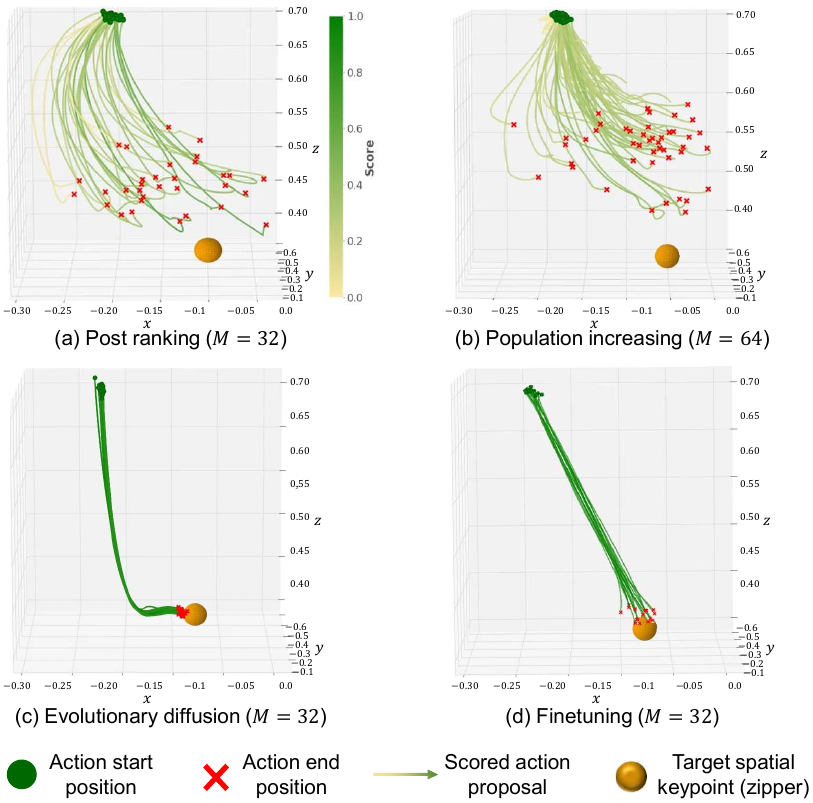}
\caption{\textbf{Visualization of left-arm action proposals in bimanual zippering tasks.} (a) Baseline steering methods relying on post-ranking and selection fail due to lack of viable initial proposals; (b) increasing the number of proposals does not guarantee feasible actions; (c) our method employs Evolutionary Diffusion to evolve action proposals toward a target distribution with higher task alignment that is comparable to (d) fine-tuned policies.}
\label{fig6}
\end{figure}
% \vspace{-15pt}

\begin{table}[!t]
\caption{Out-of-distribution task performance}
\label{tab2}
\centering
% \small
\setlength{\tabcolsep}{4pt} % 控制列内左右间距
\resizebox{\columnwidth}{!}{ % 自动缩放表格宽度至半栏宽
\begin{tabular}{@{}l@{\hskip 6pt}c@{\hskip 6pt}c@{\hskip 6pt}c@{}}
\toprule
\textbf{Tasks} & \textbf{V-GPS} & \textbf{FOREWARN} & \textbf{Ours} \\
\midrule
Basket Flipping         & 0.20 \scriptsize$\pm$ 0.03  & 0.36 \scriptsize$\pm$ 0.07 & 0.64 \scriptsize$\pm$ 0.02 \\
Bimanual Bussing     & 0.10 \scriptsize$\pm$ 0.01  & 0.13 \scriptsize$\pm$ 0.05 & 0.54 \scriptsize$\pm$ 0.03 \\
Bimanual Zippering   & 0.05 \scriptsize$\pm$ 0.02  & 0.07 \scriptsize$\pm$ 0.01 & 0.33 \scriptsize$\pm$ 0.035 \\
\midrule
\textbf{Overall}     & 0.12 & 0.19 & \textbf{0.50} \\
\bottomrule
\end{tabular}
}
\end{table}

\textbf{VLA-Pilot Outperforms Steering Baselines.}
We compare VLA-Pilot against V-GPS and FOREWARN by integrating each with DiVLA. As shown in Table~\ref{tab1}, VLA-Pilot achieves higher average MSR and SOA across tasks. On simple tasks (e.g., Mug Handling, Bag Handling), all methods perform similarly. However, VLA-Pilot shows clear advantages on complex tasks (e.g., Bimanual Bussing, Zippering). We attribute this advantage to the proposed evolutionary strategy. In simple tasks, pre-trained VLA polices typically generate candidates that already include feasible behaviors (e.g., approaching the mug handle or bag corner), making static ranking and selection sufficient. In contrast, complex tasks demand fine-grained coordination, where initial candidates often lack viable execution patterns. Baseline methods that rely solely on post-ranking and selection frequently fail under such settings (Figure \ref{fig6}(a)). Even increasing the number of proposals does not guarantee the inclusion of feasible actions (Figure \ref{fig6} (b)). VLA-Pilot addresses this limitation by employing Evolutionary Diffusion to evolve elite action proposals toward a target distribution with higher feasibility and task alignment (Figure \ref{fig6}(c)). This enables VLA-Pilot to achieve performance comparable to fine-tuned policies (Figure \ref{fig6}(d)). Under OOD settings, VLA-Pilot maintains strong performance with an average MSR of 0.50 (Table~\ref{tab2}), while V-GPS and FOREWARN drop to 0.12 and 0.19, respectively. This generalization capability stems from the use of MLLMs for open-world objective reasoning, in contrast to the dataset-specific verifiers used by baselines.

\begin{figure}[!t]
\centering
\includegraphics[width=0.95\linewidth]{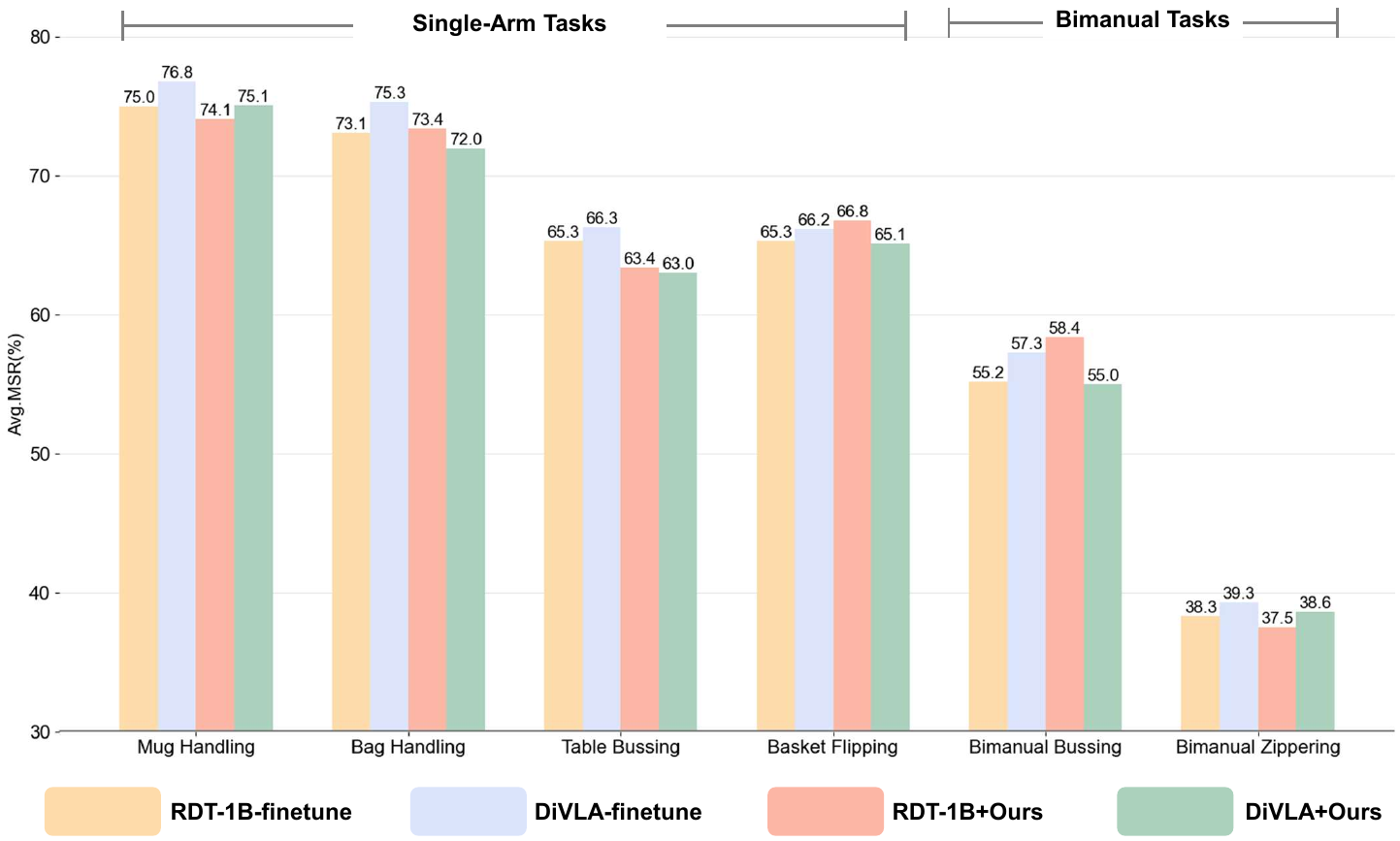}
\caption{\textbf{Comparison with VLA fine-tuning.} VLA-Pilot achieves performance comparable to VLA fine-tuning methods with 50 demonstrations.}
\label{fig7}
\end{figure}

\textbf{VLA-Pilot Matches Fine-tuning Performance.} We compare VLA-Pilot with supervised fine-tuning methods, namely DiVLA-finetune and RDT-1B-finetune on all six tasks. As shown in Figure~\ref{fig7}, VLA-Pilot achieves performance comparable to these fine-tuning approaches. Notably, these results support a key finding from earlier experiments: the failures of pre-trained VLA policies are often not due to skill inefficiency, but rather task misperception or suboptimal action selection. In such cases, fine-tuning the policy may be unnecessary. Instead, adapting existing policies via inference-time steering can effectively recover their capabilities. 

\begin{figure}[!t]
\centering
\includegraphics[width=0.95\linewidth]{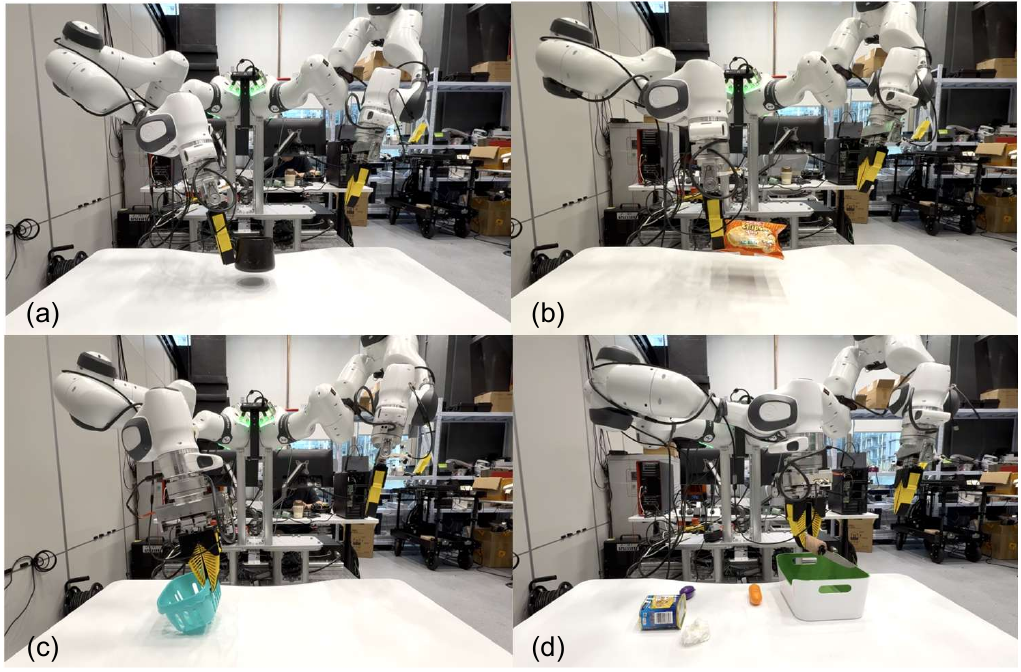}
\caption{\textbf{Qualitative results of cross-embodiment experiments.} VLA-Pilot achieves zero-shot generalization on the Franka robot, maintaining consistent task performance across four single-arm tasks.}
\label{fig8}
\end{figure}

\begin{table}[htbp]
\caption{Cross-embodiment task performance}
\label{tab3}
\centering
% \small
\setlength{\tabcolsep}{4pt} % 控制列内左右间距
\resizebox{\columnwidth}{!}{ % 自动缩放表格宽度
\begin{tabular}{@{}l@{\hskip 6pt}c@{\hskip 6pt}c@{\hskip 6pt}c@{\hskip 6pt}c@{}}
\toprule
\textbf{Method} & \textbf{\makecell{Mug \\ Handling}} & \textbf{\makecell{Bag \\ Handling}} & \textbf{\makecell{Basket \\ Flipping}} & \textbf{\makecell{Table \\ Bussing}} \\
\midrule
DiVLA & 0.55 \scriptsize$\pm$ 0.03 & 0.54 \scriptsize$\pm$ 0.07 & 0.45 \scriptsize$\pm$ 0.04 & 0.25 \scriptsize$\pm$ 0.02 \\
DiVLA+Ours & 0.78 \scriptsize$\pm$ 0.02 & 0.75 \scriptsize$\pm$ 0.03 & 0.67 \scriptsize$\pm$ 0.05 & 0.56 \scriptsize$\pm$ 0.04 \\
\midrule
\textbf{Improvement} & +0.23 & +0.21 & +0.22 & +0.31 \\
\bottomrule
\end{tabular}
}
\end{table}

\textbf{VLA-Pilot Achieves Cross-Embodiment Generalization.}
We deploy VLA-Pilot on the Franka Panda robot in a zero-shot setting. Compared to the pre-trained DiVLA baseline, VLA-Pilot achieves notable performance gains across all tasks (see Table~\ref{tab3}):
+0.23 in Mug Handling, +0.21 in Bag Handling, +0.22 in Basket Flipping, and +0.31 in Table Bussing. These results demonstrate robust, consistent behavior across embodiments, as further illustrated in Figure~\ref{fig8}.

\begin{figure}[!t]
\centering
\includegraphics[width=0.95\linewidth]{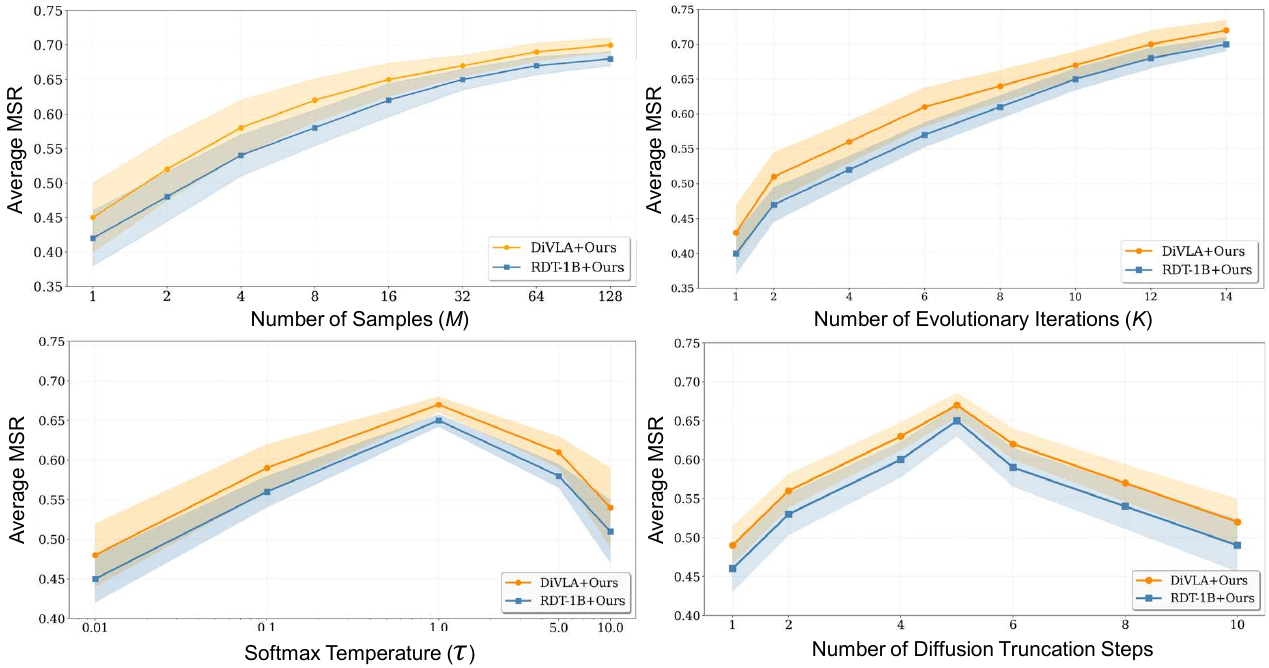}
\caption{\textbf{Hyperparameter sensitivity analysis of VLA-Pilot.} Performance improves with more samples \(M\) and evolutionary steps \(K\), which aligns with the inference-time scaling law \cite{kwok2025robomonkey}. Optimal performance occurs at truncation steps = 5 and temperature \(\tau{=}1.0\).} 
\label{fig9}
\end{figure}

\textbf{Ablation Study.} 
We ablate each core module of VLA-Pilot in simulation (Table~\ref{tab:sim_results}). Removing EPS-CoT (replaced by heuristic prompting) results in an 8\% MSR drop, highlighting the need for multimodal reasoning and embodied grounding. Replacing Evolutionary Diffusion with selection-only reduces MSR by 11\%, confirming its role in adaptive action optimization. Removing Iterative Steering Refinement causes 5\% MSR drop, due to cascading errors in open-loop execution. Together, these results show all modules are essential and complementary to VLA-Pilot's robustness. We also analyze hyperparameter sensitivity for four key values: action samples $M$, evolution steps $K$, diffusion truncation steps, and softmax temperature $\tau$. As shown in Figure~\ref{fig9}, larger $M$ and $K$ improve MSR. Truncation steps beyond 5 degrade performance due to over-exploration. Optimal performance is achieved with $\tau=
1.0$, balancing action diversity and convergence.

\textbf{Inference-time Latency Analysis.} 
We report average per-step latency across modules in Table~\ref{tab:time-efficiency}, with overall inference time at 2.41s. The main bottleneck lies in MLLM-based reward reasoning. This latency reflects a reasonable trade-off between generalization and efficiency, similar to prior FMs-based steering pipelines like FOREWARN (3.7s)~\cite{wu2502foresight}. Future work may reduce latency via local deployment and model pruning~\cite{dingmllm}.

\begin{table}[htbp]
\centering
\caption{Inference Time Latency of VLA-Pilot}
\label{tab:time-efficiency}
\begin{tabular}{lc}
\toprule
\textbf{Module} & \textbf{Latency (s) / Steering Step} \\
\midrule
EPS-CoT & $0.72 \pm 0.03$ \\
VLA sampling & $0.41 \pm 0.02$ \\
Evolutionary Diffusion & $0.52 \pm 0.01$ \\
Iterative Refinement & $0.76 \pm 0.02$ \\
\midrule
\textbf{Overall} & $2.41 \pm 0.02$ \\
\bottomrule
\end{tabular}
\end{table}

\subsection{Discussion}
VLA-Pilot enables effective inference-time steering of pre-trained VLA policies without fine-tuning. Compared to prior methods, it offers three key advantages. First, it uses MLLMs as open-world verifiers, removing the need for task-specific training and improving generalization to OOD tasks (Table~\ref{tab2}). Second, Evolutionary Diffusion adaptively refines action candidates, overcoming the limitations of static selection (Table~\ref{tab1}). Third, closed-loop refinement allows reactive correction under dynamic conditions. Notably, VLA-Pilot matches the performance of supervised fine-tuning with 50 expert demonstrations (Figure~\ref{fig7}), highlighting its practical value as a lightweight and generalizable alternative for real-world VLA deployment.
\section{Conclusions}
In this paper, we presented VLA-Pilot, an inference-time policy steering method that enables zero-shot deployment of pre-trained VLA models without any fine-tuning. Both simulation and real-world experiments validate its effectiveness and highlight its potential as a universal and modular plug-in for aligning generalist VLA policies with diverse downstream task goals. Despite its effectiveness, VLA-Pilot has several limitations. First, it assumes that the underlying VLA policy supports noise-conditioned sampling, limiting its applicability to diffusion-based architectures. Second, the reliance on MLLMs introduces non-trivial inference-time latency. Third, the current reward evaluation paradigm depends on keypoint-based vision grounding, which may be brittle in tasks involving delayed effects or deformable object interactions. Future directions include extending the steering paradigm to broader VLA architectures, optimizing MLLM integration via quantization or caching strategies, and improving robustness by incorporating richer embodied feedback beyond keypoint-level signals.

\bibliographystyle{IEEEtran}
\bibliography{references}

\end{document}